\documentclass[10pt,letterpaper]{article}

\usepackage[margin=0.75in]{geometry}
\usepackage{graphicx}
\usepackage{booktabs}
\usepackage{array}
\usepackage{tabularx}
\usepackage{amssymb}
\usepackage{hyperref}
\usepackage{xcolor}
\usepackage{enumitem}
\usepackage{caption}
\usepackage{subcaption}
\usepackage{float}
\usepackage{natbib}

\hypersetup{
  colorlinks=true,
  linkcolor=blue!50!black,
  citecolor=blue!50!black,
  urlcolor=blue!50!black
}

\setlist[itemize]{leftmargin=1.25em, itemsep=0.2em, topsep=0.25em}
\title{\vspace{-1.5em}RoboMME-Interference: Benchmarking Robot Memory Under Interference}
\author{Soumil Rathi\\\normalsize Independent Researcher\\[3pt]}
\date{August 2026}

\begin{document}
\maketitle

\begin{abstract}
Robots deployed in realistic settings will accumulate experience across many sessions and tasks over their deployment. The robot's tasks may often require it to remember information from multiple sessions ago, making long-context robot memory important for real-world deployments. However, most robot-memory benchmarks today are based on single episodes or a short context. To measure how current robot memory systems perform on longer sessions with more distractions, we introduce \textbf{RoboMME-Interference}, a cross-session benchmark built on RoboMME~\citep{dai2026robomme}. For each query episode, we construct a session history using the query's relevant prior demonstration followed by a controlled number of unrelated sessions, which we provide to the VLA as memory and measure accuracy. Running RoboMME's released memory-augmented $\pi_{0.5}$ variants unmodified through this benchmark, we find that while perceptual memory variants improve success when given the history without any distractors, they decay strongly and steadily as unrelated sessions accumulate. The subgoal variants, which read the history with a vision-language model and pass written subgoals to the policy, improve less at their best but hold more of that improvement as distractors accumulate. Adding a retrieval step to the strongest perceptual variant, which selects the section of history most visually similar to the robot's current view and passes only that section to the policy, restores its no-distractor success rate at every interference level. With this release, we emphasize the importance of long-context memory and robustness to interference and show that current systems largely fail on such capabilities. The project page, videos, code, and data are at \url{https://robotmemorybench.com}.

\end{abstract}

\section{Introduction}

As robots are increasingly deployed in real-world environments~\citep{black2024pi0, physicalintelligence2025pi05}, they will quickly need to operate across many sessions and many unrelated tasks. The experience or information that matters for the current task may sit far back in that history, behind sessions that have nothing to do with it. Thus, any robot memory system must be capable of working with long-context information recall and robust to interference.

The core question is: when a useful prior session is embedded in a longer history of unrelated sessions, are current robot-memory systems still able to recall it? We answer it with RoboMME-Interference, a cross-session benchmark built on RoboMME~\citep{dai2026robomme} that inserts a controlled number of unrelated sessions between a relevant memory and the query.

\paragraph{Contributions.}
\begin{itemize}
  \item \textbf{A cross-session interference benchmark.} We extend RoboMME by adding $k \in \{0,1,3,7\}$ unrelated sessions (from different tasks) to each task and measuring each memory system's accuracy, allowing us to measure the impact of session length and additional unrelated memory on accuracy directly.
  \item \textbf{A complete, released result grid.} We showcase the performance of twelve RoboMME memory systems on all nine eligible task families, showing the impact on each memory system and task type individually, released alongside the code, data, figures, and analysis.
  \item \textbf{The decay is recoverable.} We add a retrieval step to FrameSamp-Modul that finds the demonstration by visual similarity and passes only it to the policy; its success rate stays between 43.6\% and 44.7\% at every interference level, where the unmodified variant falls from 45.3\% to 19.3\%.
\end{itemize}

\section{Related Work}

Recent VLA policies~\citep{brohan2023rt2, kim2024openvla, black2024pi0} add memory in functionally distinct ways: some retrieve relevant past experience on demand~\citep{li2025mapvla, sridhar2025memer}, some keep a store of past observations the policy reads from~\citep{shi2025memoryvla, guo2026chameleon}, and some summarize recent experience into a compact state~\citep{torne2026mem}. 

Language models face the same problem, and the LLM community measures it directly: multi-session benchmarks place the information a query needs inside a long history of unrelated conversation and test whether the model can still use it~\citep{maharana2024locomo, wu2025longmemeval, tavakoli2025beam}. Recall degrades as the unrelated history grows~\citep{liu2024lostinthemiddle}. The memory tools built in response, retrieval~\citep{lewis2020rag}, recurrence, and compression, are the same systems now appearing in robot policies.

Robot-memory benchmarks~\citep{dai2026robomme, chen2026rmbench, lei2026robomemarena} measure whether a policy can use memory within a single episode. None measure whether that memory holds up as the session history grows and fills with unrelated sessions, the failure the LLM benchmarks document. We add that axis to RoboMME, reusing its tasks, variants, and released checkpoints.

\section{The Benchmark}

\begin{figure}[H]
  \centering
  \includegraphics[width=\linewidth]{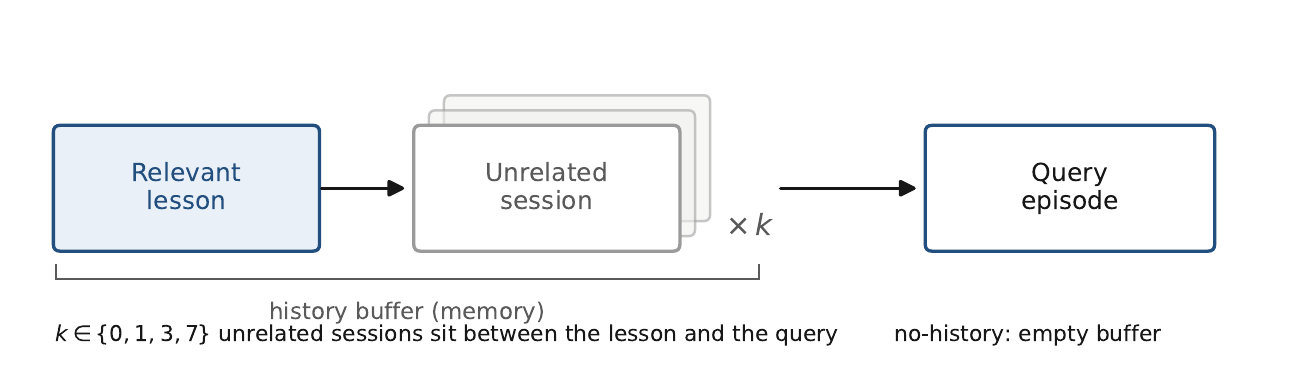}
  \caption{The history buffer: the relevant lesson, then $k$ unrelated sessions, then the query. Larger $k$ pushes the lesson farther back.}
  \label{fig:protocol}
\end{figure}

\paragraph{The history buffer.}
Each query episode links to an external history buffer. This history buffer, or memory, contains the query episode's original demonstration section (the evidence the task needs), which is followed by $k$ unrelated distractor sessions. Observations from the current episode also accumulate in this buffer during the rollout. We sweep $k \in \{0,1,3,7\}$ (none, light, moderate, and strong interference) against a \texttt{no-history} baseline with an empty buffer.

Each distractor contributes one fixed-length section of unrelated history so that our $k$ value is comparable across different distractor sessions. We set each distractor unit to 32 stored frames, matching the number of frames that FrameSamp, the core perceptual variant, uses after sampling the whole history. One distractor therefore contains as much material as the policy's entire sampled view.\footnote{Each distractor unit is the first 256 raw frames of the distractor episode subsampled at stride 8.} Every system consumes the buffer through its own unmodified mechanism (FrameSamp samples frames, TokenDrop selects tokens, recurrent variants process sampled history), with no per-condition retraining; only the buffer contents change. The $\pi_{0.5}$ baseline has no memory and appears only at \texttt{no-history}.

\paragraph{Which tasks, and why.}
Our benchmark requires the prior relevant information to be in a self-contained section, so that it can be cleanly split and added to the external history buffer with distractors. We therefore use the nine of RoboMME's sixteen tasks~\citep{dai2026robomme} that deliver this information as a prior demonstration video, distinct from the execution section: the four imitation tasks (MoveCube, InsertPeg, PatternLock, RouteStick), the three tasks where the demonstration shows which object or target to choose (VideoRepick, VideoPlaceButton, VideoPlaceOrder), and the two tasks where it shows where a hidden cube is located (VideoUnmask, VideoUnmaskSwap). The remaining seven interleave their cue inside a single execution episode: the counting suite accumulates action counts during execution, and the button- and highlight-based variants (ButtonUnmask, ButtonUnmaskSwap, PickHighlight) reveal their cue mid-episode. They therefore offer no separable prior section to place under cross-session interference. The final nine tasks we chose are shown in Table~\ref{tab:tasks}.

\begin{table}[H]
  \centering
  \small
  \begin{tabularx}{\linewidth}{llX}
    \toprule
    Task & Memory type & What the prior session must supply \\
    \midrule
    VideoRepick      & Object     & Which object was picked up in the demonstration, to re-pick it. \\
    VideoPlaceOrder  & Object     & Which target in an ordered set the cube belongs on. \\
    VideoPlaceButton & Object     & The cube-to-target placement shown in the demonstration. \\
    VideoUnmask      & Spatial    & Where an occluded cube of a given color is located. \\
    VideoUnmaskSwap  & Spatial    & The occluded-cube location after the cubes have been swapped. \\
    MoveCube         & Procedural & The demonstrated manner of moving the cube to its target. \\
    InsertPeg        & Procedural & Which peg to grasp and how to insert it, from the demonstration. \\
    PatternLock      & Procedural & A continuous pattern traced over a grid, to be retraced. \\
    RouteStick       & Procedural & A path through obstacles, including turn directions, to repeat. \\
    \bottomrule
  \end{tabularx}
  \caption{The nine task families, all drawn from RoboMME~\citep{dai2026robomme}. Each presents a prior demonstration the policy must use during a separate execution episode.}
  \label{tab:tasks}
\end{table}

\paragraph{Why different-family distractors.}
We draw distractors from \emph{different} task families than the query. A distractor from the same task family could assert a conflicting fact (say, a different target cube or a contradictory sequence to copy), thus forcing the VLA to resolve contradictions and making any degradation uninterpretable. Instead, cross-family distractors are simply irrelevant, allowing us to scale interference without introducing contradictory information.

\section{Memory Systems}

We run RoboMME's released memory-augmented variants through our benchmark. Each variant pairs a memory representation (e.g. frames, tokens, hidden states) with a mechanism for how that memory is integrated with the base $\pi_{0.5}$ policy~\citep{physicalintelligence2025pi05}. 

\textbf{Context} concatenates memory tokens with the input, \textbf{Modulator} conditions the policy via adaptive LayerNorm, and \textbf{Expert} routes memory through a dedicated expert using block-wise causal attention~\citep{dai2026robomme}. The perceptual representations reduce or resample visual-history tokens~\citep{bolya2023tome, yu2023sevila}; the recurrent representation maintains a compressed hidden state via test-time training~\citep{sun2024ttt}, following recurrent memory transformers~\citep{bulatov2022rmt}. In the symbolic variants, a vision-language model passes written subgoals to the policy, as plain text (SimpleSG) or with image coordinates for the objects it references (GroundSG). SimpleSG and GroundSG receive up to 10 frames sampled across the demonstration and distractor video. They also receive the current image and previous predicted subgoals. MemER~\citep{sridhar2025memer} receives the same video, recent execution frames, and keyframes it selects during execution.

We exclude the recurrent variants without released checkpoints as of this time (all RMT, and TTT-Modulator).

\begin{table}[H]
  \centering
  \small
  \begin{tabular}{llccc}
    \toprule
    Representation & Type & Context & Modulator & Expert \\
    \midrule
    FrameSamp & perceptual & \checkmark & \checkmark & \checkmark \\
    TokenDrop & perceptual & \checkmark & \checkmark & \checkmark \\
    TTT       & recurrent  & \checkmark & --         & \checkmark \\
    RMT       & recurrent  & --         & --         & -- \\
    Symbolic  & SimpleSG / GroundSG & \multicolumn{3}{c}{\checkmark\ (a VLM reads the buffer and writes subgoals)} \\
    MemER     & keyframe + subgoal & \multicolumn{3}{c}{\checkmark\ (subgoals, with selected key frames kept)} \\
    \bottomrule
  \end{tabular}
  \caption{The system grid. \checkmark{} marks the eleven memory variants we evaluate; with the $\pi_{0.5}$ baseline, twelve systems.}
  \label{tab:systems}
\end{table}

\section{Results}

The complete grid contains twelve systems evaluated on nine task families (each with all 50 test episodes), with each memory system evaluated across five total session lengths (and the baseline evaluated with no memory). We report confidence intervals (Wilson 95\%) for each success rate since each rate is estimated from a finite set of episodes. 

\begin{table}[H]
  \centering
  \small
  \begin{tabularx}{\linewidth}{lccccc}
    \toprule
    System & No history & k0 & k1 & k3 & k7 \\
    \midrule
    $\pi_{0.5}$ (baseline) & 17.3\% & -- & -- & -- & -- \\
    FrameSamp-Modul        & 18.2\% & \textbf{45.3\%} & \textbf{38.4\%} & \textbf{30.0\%} & 19.3\% \\
    TokenDrop-Modul        & 17.1\% & 35.3\% & 31.3\% & 24.2\% & 20.7\% \\
    FrameSamp-Context      & 17.8\% & 26.7\% & 23.8\% & 20.0\% & 17.3\% \\
    FrameSamp-Expert       & 17.6\% & 27.1\% & 27.3\% & 20.0\% & 19.3\% \\
    TokenDrop-Context      & 13.8\% & 22.9\% & 21.1\% & 16.2\% & 14.2\% \\
    TokenDrop-Expert       & 16.9\% & 27.3\% & 22.2\% & 17.3\% & 15.6\% \\
    Recurrent-TTT-Expert   & 16.0\% & 18.0\% & 16.4\% & 14.9\% & 16.4\% \\
    Recurrent-TTT-Context  & 16.9\% & 15.3\% & 16.4\% & 17.3\% & 13.8\% \\
    SimpleSG               & 18.0\% & 24.2\% & 22.7\% & 20.9\% & 21.8\% \\
    GroundSG               & \textbf{20.7\%} & 34.2\% & 32.0\% & 26.4\% & \textbf{26.0\%} \\
    MemER                  & 18.2\% & 32.2\% & 27.8\% & 22.9\% & 21.3\% \\
    \bottomrule
  \end{tabularx}
  \caption{Success rates across all nine task families.}
  \label{tab:main-success}
\end{table}

\begin{figure}[H]
  \centering
  \includegraphics[width=\linewidth]{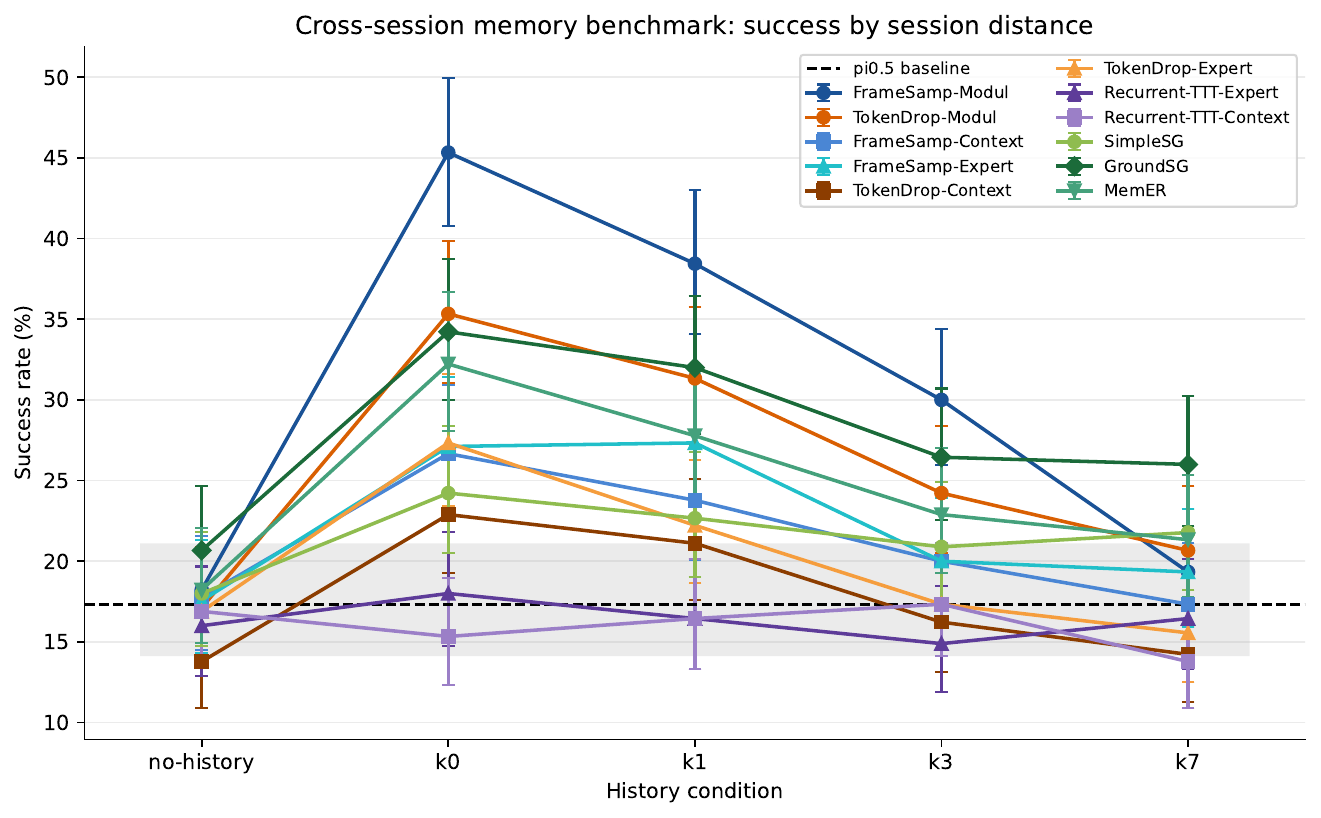}
  \caption{Overall success by memory system and history condition across all nine families. The gain at $k_0$ is largest for the perceptual modulator variants but falls back toward the no-history floor as distractors are added.}
  \label{fig:overall}
\end{figure}

With only the relevant lesson in the buffer ($k_0$), as RoboMME also found, perceptual memory lifts accuracy far above each system's own no-history floor: FrameSamp-Modul, the clear winner, boosts accuracy by $+27.1$\,pp (from 18.2\% to 45.3\%). However, as unrelated sessions accumulate, that gain erodes steadily: FrameSamp-Modul falls $\sim$26\,pp from $k_0$ to $k_7$ and TokenDrop-Modul $\sim$15\,pp, leaving both within a few points of their no-history floors.

Two systems with the same representation can behave very differently depending on how memory enters the policy. With the same FrameSamp representation, the Modulator variant reaches 45.3\% at $k_0$ versus 26.7\% (Context) and 27.1\% (Expert), and TokenDrop shows the same pattern (35.3\% vs.\ 22.9\% and 27.3\%). The recurrent TTT variants extract no meaningful gain at any session length, staying at or below 18\% at $k_0$, essentially their no-history level.

The three subgoal variants trade peak benefit for slower decay. Their gains at $k_0$ are smaller than FrameSamp-Modul's, but so are their losses as distractors accumulate: GroundSG falls 8.2\,pp from $k_0$ to $k_7$ where FrameSamp-Modul falls 26.0\,pp, and at $k_7$ GroundSG's 26.0\% is the highest rate of any system. It also ends above its own no-history floor at $k_7$, where FrameSamp-Modul ends within noise of its floor.

\begin{figure}[H]
  \centering
  \includegraphics[width=\linewidth]{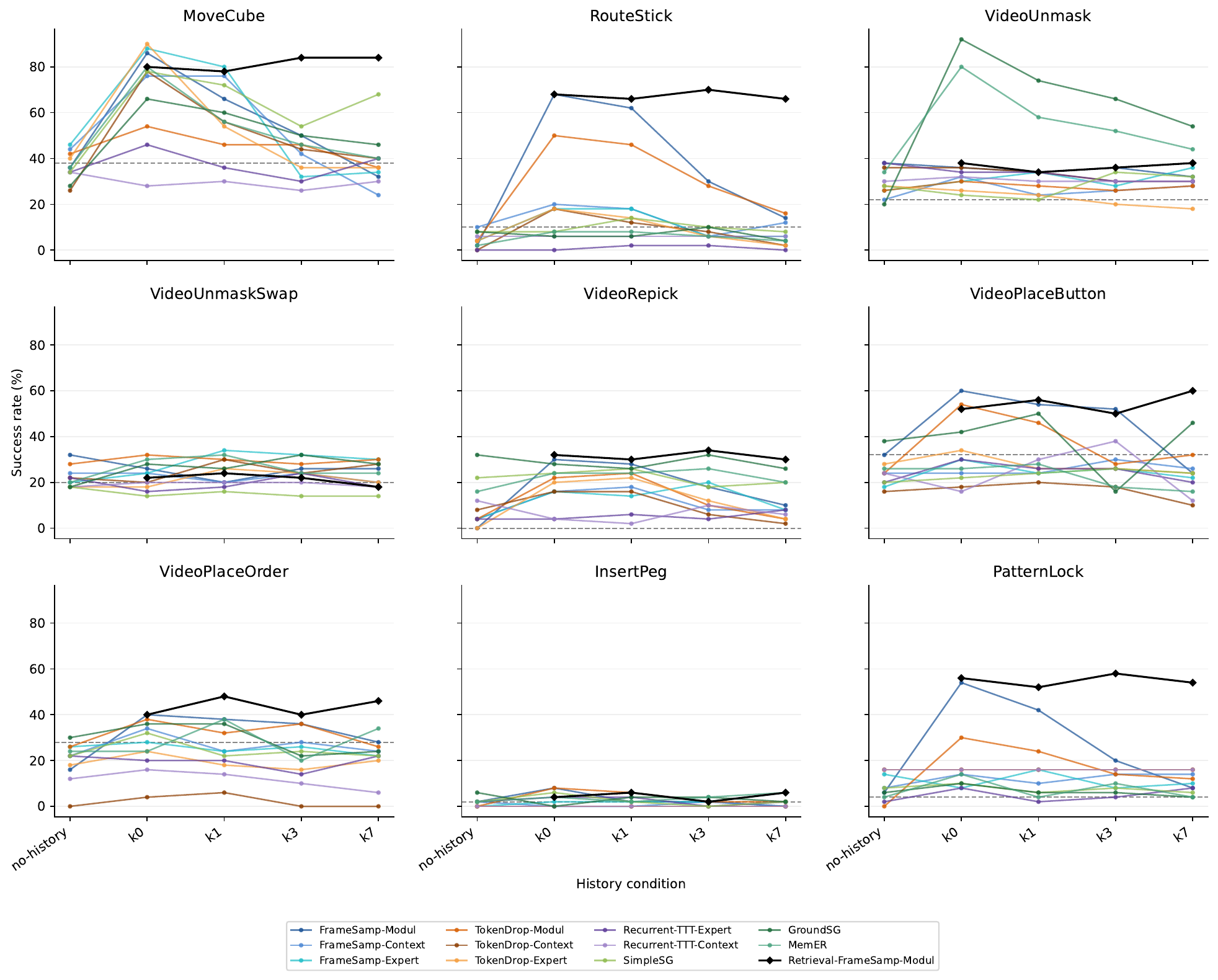}
  \caption{Per-family success curves for all evaluated systems, with Retrieval-FrameSamp-Modul (black; Section~\ref{sec:retrieval}) shown for reference. Families differ in floors, ceilings, and memory sensitivity.}
  \label{fig:family-curves}
\end{figure}

\begin{figure}[H]
  \centering
  \includegraphics[width=0.86\linewidth]{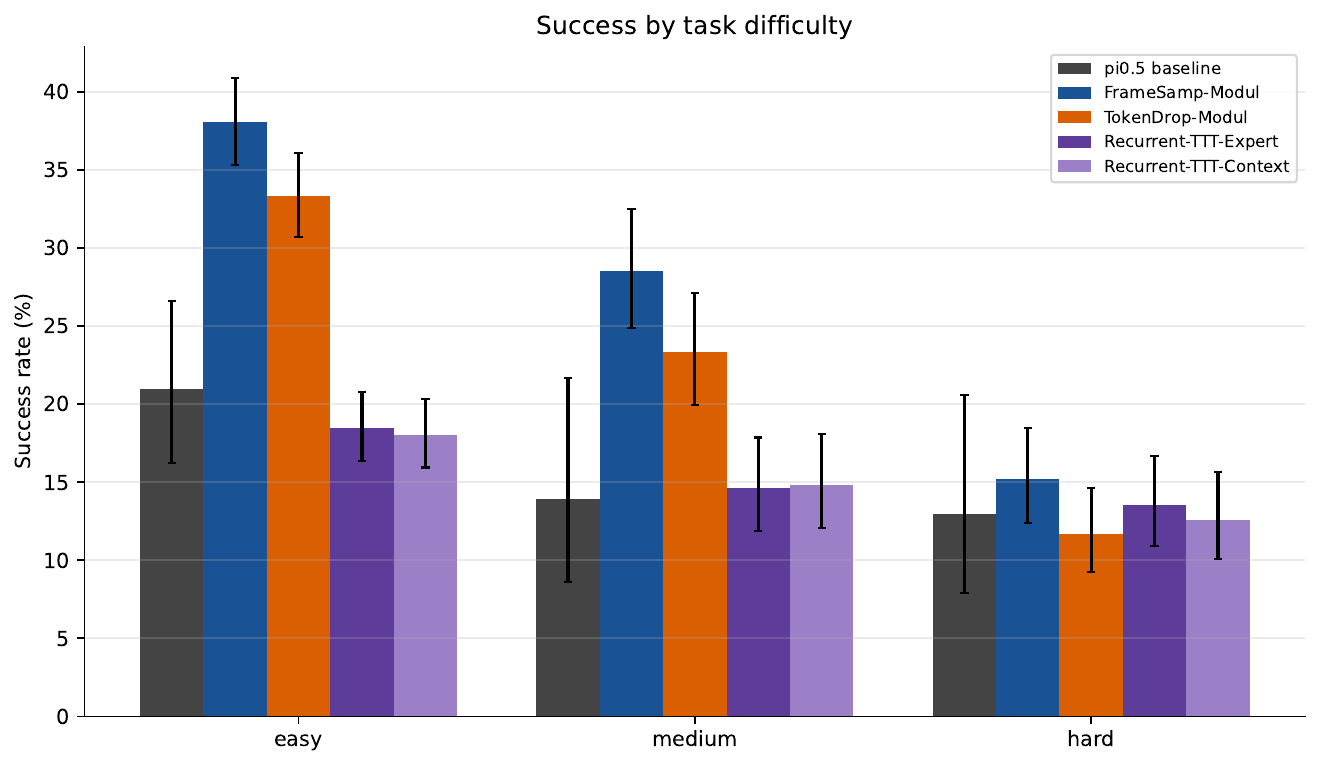}
  \caption{Success by RoboMME's easy/medium/hard difficulty stratification.}
  \label{fig:difficulty}
\end{figure}

The effect is also uneven across task family and task difficulty. The differences in memory effect per task family indicate how some tasks may be better helped by memory or are too difficult for the policy to solve even with memory (Figure~\ref{fig:family-curves}). We see differences per difficulty as well: the lift is largest on RoboMME's easy and medium episodes, while hard episodes stay near the floor for every memory system (Figure~\ref{fig:difficulty}).

\section{Adding Retrieval}\label{sec:retrieval}

To test whether the decay is recoverable, we add a retrieval step in front of FrameSamp-Modul, the strongest variant. The step embeds every frame with SigLIP~\citep{zhai2023siglip}, splits the history into sessions wherever the similarity between adjacent frames drops sharply (a visual scene change), and passes the session that best matches the robot's current observation to the policy.\footnote{Frames within a session are far more similar than frames across a session cut. We chose 0.923 as the threshold, which lies in the gap between the two; any value from 0.90 to 0.95 gave similar results.} To match, we embed the current observation and select the session containing its closest frame. The policy and encoder are used as released, and the benchmark's ground-truth session boundaries are never used directly.

\begin{table}[H]
  \centering
  \small
  \begin{tabular}{lcccc}
    \toprule
    System & k0 & k1 & k3 & k7 \\
    \midrule
    FrameSamp-Modul           & 45.3\% & 38.4\% & 30.0\% & 19.3\% \\
    Retrieval-FrameSamp-Modul & 43.6\% & 43.8\% & 44.0\% & 44.7\% \\
    \bottomrule
  \end{tabular}
  \caption{Success rates across all nine task families, with and without the retrieval step.}
  \label{tab:retrieval}
\end{table}

\begin{figure}[H]
  \centering
  \includegraphics[width=\linewidth]{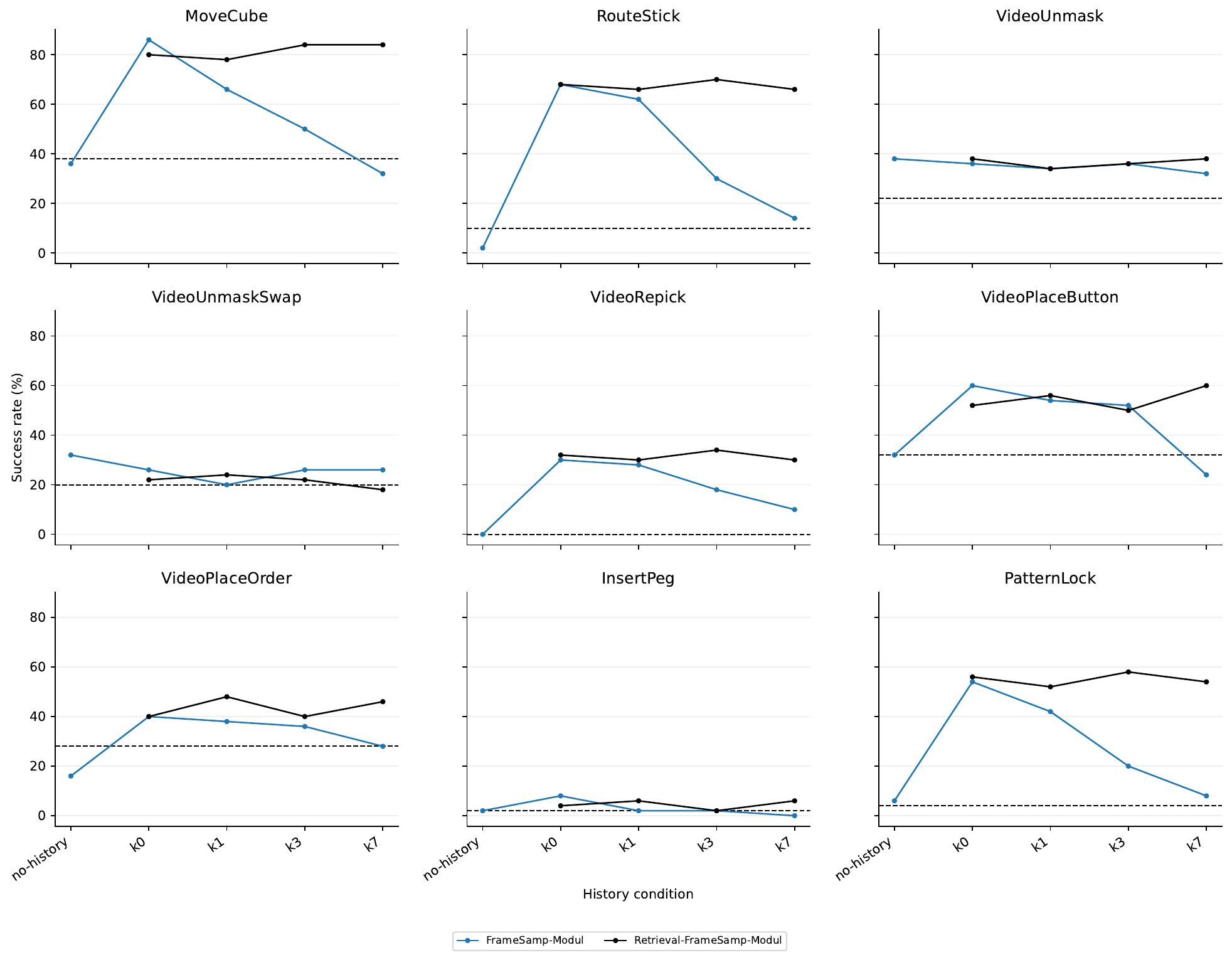}
  \caption{Per-family success for FrameSamp-Modul with and without the retrieval step. The dashed line is the $\pi_{0.5}$ \texttt{no-history} baseline.}
  \label{fig:retrieval-perfamily}
\end{figure}

Retrieval-FrameSamp-Modul restores the $k_0$ success rate through $k_7$ on seven of the nine families (Figure~\ref{fig:retrieval-perfamily}). Across all nine, it stays between 43.6\% and 44.7\% at every interference level, where FrameSamp-Modul falls from 45.3\% to 19.3\% (Table~\ref{tab:retrieval}).

On the remaining two families, VideoUnmask and VideoUnmaskSwap, memory does not help in the first place: FrameSamp-Modul's $k_0$ success rate is no higher than its \texttt{no-history} rate, so there is no benefit to lose and none to recover. Retrieval-FrameSamp-Modul performs comparably there.

The recovery suggests that the policy can use the demonstration, but its memory system struggles to pick out the relevant frames from unrelated sessions without retrieval. Retrieval is easy in this setup. The robot's current view looks like the demonstration video, the demonstration is always somewhere in the buffer, and the buffer splits cleanly into sessions at sharp scene changes rather than arriving as one continuous stream.

\section{Limitations}

\begin{itemize}
  \item This benchmark runs in simulation, on RoboMME's tasks and released checkpoints; testing on real robots is future work.
  \item We use one checkpoint per system and report one rollout per episode and condition, so the results capture variation across episodes but not across training runs or repeated rollouts.
  \item Only nine of RoboMME's sixteen tasks provide a demonstration that can be separated from the episode and placed in the history buffer, so the other seven cannot be tested this way.
  \item The released data reports each episode's outcome and step count, not full step-by-step traces.
  \item The retrieval result has its own caveats: we test it only on FrameSamp-Modul, and it relies on the current view looking like the demonstration, which real robot histories may not guarantee.
\end{itemize}

\section{Conclusion}

Memory helps when the relevant session is close, but most systems we test lose most of that benefit as the history grows: the perceptual and recurrent variants return to roughly their no-history performance once enough unrelated sessions intervene, while the subgoal variants keep a small advantage at the cost of a lower peak. A simple retrieval step in front of the strongest perceptual variant restores its full $k_0$ gain at every interference level, though only because the query resembles its demonstration here.

Deployed histories will often be harder: the experience may arrive as one continuous stream with no clean session boundaries, the relevant session may share the task but not its appearance, and the needed information may be spread across several sessions. Cross-session recall under those conditions remains open. It is, however, the same problem the LLM memory literature has worked on for years, thus providing a natural body of work to build on as robot memory matures.

\bibliographystyle{plainnat}
\bibliography{references}

\end{document}